\documentclass[tinyml]{acmart}

\AtBeginDocument{%
  \providecommand\BibTeX{{%
    \normalfont B\kern-0.5em{\scshape i\kern-0.25em b}\kern-0.8em\TeX}}}

\setcopyright{rightsretained}
\copyrightyear{2022}
\acmYear{2022}
\usepackage{subcaption}
\usepackage{multirow}
\usepackage{booktabs}

\begin{document}

\title{IMU Preintegrated Features for Efficient Deep Inertial Odometry}


\author{Rooholla Khorrambakht}
\affiliation{%
  \institution{K.N.Toosi University of Technology}
  \city{Tehran}
  \country{Iran}
}
\email{r.khorrambakht@email.kntu.ac.ir}

\author{Hamed Damirchi}
\affiliation{%
  \institution{K.N.Toosi University of Technology}
  \city{Tehran}
  \country{Iran}}
\email{hdamirchi@email.kntu.ac.ir}

\author{Hamid D. Taghirad}
\affiliation{%
  \institution{K.N.Toosi University of Technology}
  \city{Tehran}
  \country{Iran}
}
\email{taghirad@kntu.ac.ir}



\begin{abstract}
MEMS Inertial Measurement Units (IMUs) as ubiquitous proprioceptive motion measurement devices are available on various everyday gadgets and robotic platforms. Nevertheless, the direct inference of geometrical transformations or odometry based on these data alone is a challenging task. This is due to the hard-to-model imperfections and high noise characteristics of the sensor, which has motivated research in formulating the system as an end-to-end learning problem, where the motion patterns of the agent are exploited to facilitate better odometry estimates. However, this benefit comes at the cost of high computation and memory requirements, which makes deep inertial odometry unsuitable for low-power and edge applications. This paper attempts to address this conflict by proposing the IMU preintegrated features as a replacement for the raw IMU data in deep inertial odometry. Exploiting the manifold structure of the IMU motion model, these features provide a temporally compressed motion representation that preserves important geometrical information. We demonstrate the effectiveness and efficiency of this approach for the task of inertial odometry on two applications of pedestrian motion estimation and autonomous vehicles. We show a performance improvement compared to raw inputs while reducing the computational burdens. Additionally, we demonstrate the efficiency of this approach through an embedded implementation on a resource-constrained microcontroller.  
\end{abstract}


\maketitle
\section{Introduction}
\label{introduction}
With the advent of MEMS Inertial Measurement Units (IMUs) as low-power and low-cost proprioceptive motion sensors, this modality has been added to numerous everyday gadgets
and autonomous systems. Smartphones and smartwatches, autonomous vehicles, VR\footnote{Virtual Reality} headsets, and even gaming consoles all contain some variant of this sensor. This widespread availability of this modality has encouraged new applications such as user activity recognition \cite{damirchi2020arc} and pedestrian tracking \cite{Chen2020} for situations where other sensors may fail.

However, the expense of their lower cost and smaller form factor, is that MEMS IMUs are much more challenging to model accurately due to the manufacturing imperfections, high and time-variant noise characteristics, and nonlinear effects. While many prominent classical pipelines based on statistical sensor fusion and optimization systems already exist for the efficient incorporation of IMUs into various applications \cite{Sandy2019,Huang2019}, there are many scenarios where the restricting assumptions of the system's formulation limit the application and robustness of the method.

These challenges have encouraged researchers to consider deep learning methods to enable previously implausible applications of this modality as independent sensors. The ability to learn high-dimensional latent presentations from data enables these methods to exploit the underlying motion patterns of the agent to reduce the odometry drift.

Generally, there are two lines of research in this field. Many works
try to exploit leaned models as modules within a classical
fusion system to detect important motion events. For instance, \cite{Wagstaff2020} proposes using an LSTM to detect zero-velocity events in human walking to reset the states of an Extended Kalman Filter (EKF) and avoid the accumulation of error. In a similar attempt, \cite{Brossard2019} designs a classifier that detects specific motion modes of a wheeled vehicle and uses the classifier's results
to define a set of pseudo measurements for an Invariant
Extended Kalman Filter.

On the other hand, the opposing approach is to cast the problem as a
pure data-driven pipeline and synthesize models that directly
consume raw sensor data to yield the final prediction. For instance, \cite{Chen2018} proposes using an LSTM model for predicting the human trajectory represented in polar coordinates. Furthermore, this end-to-end application of IMUs has also been considered alongside other modalities such as vision \cite{Chen2019,Almalioglu2019}, and thermal images \cite{Saputra2020} to estimate the ego-motion of a robot or autonomous vehicle.

While the former approaches are computationally efficient due to their classical core and minimal models, they are highly limited to the design domain for which they are engineered. On the other hand, even though the data-driven methods are highly flexible and general, they lack the computational efficiency of their classical counterparts, and they may not be easily used for low-power embedded and wearable applications.

This paper aims to address this conflict by proposing an efficient
and temporally compact motion representation that draws inspiration from the preintegration theory known to the graph-based Visual-Inertial
Odometry (VIO) community \cite{Forster2017}. While \cite{Forster2017} formulates the preintegration as part of a classical VIO optimization problem and as a motion constraint, the novelty of this approach is in the repurposing of this geometrical constraint as a preprocessing/input-feature for learning-based models. This input feature replaces the raw IMU data and in doing so, improves the performance while extensively reducing the computational burdens. Finally, the efficiency of this approach has been further illustrated by providing an embedded implementation of a deep pedestrian inertial odometry pipeline on a microcontroller with highly restricted resources.

The rest of this paper is organized as follows. Section \ref{methods}
introduces the Preitegrated features and describes the baseline models and the training procedure used to demonstrate the effectiveness of the proposed approach. Then, Section \ref{experiments} presents the experimental setups and the results of applying preintegration for the two tasks of pedestrian and autonomous vehicles inertial odometry. Furthermore, this section also demonstrates the computational efficiency of the approach through the embedded implementation of IO-Net \cite{Chen2020} with preintegrated IMU inputs, and finally, Section \ref{conclusions} concludes the paper and proposes future directions.
\section{Methodology}
\label{methods}
This section introduces the methodology for extraction of preintegrated features from IMU raw data and presents the baseline architecture and training procedure used to demonstrate their application in deep inertial odometry. 

\begin{figure}[t]
	\centering
	\includegraphics[scale=1.25]{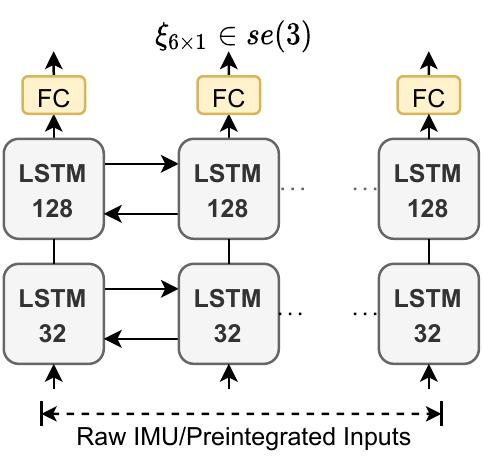}
	\caption{The baseline model for investigating the effectiveness of preintegrated features.}
	\label{base_model}
\end{figure}

\subsection{IMU Preintegrated Features}
Preintegrated IMU features draw inspiration from the factor graph approach to Visual Inertial Odometry (VIO) \cite{Forster2017}. Factor graphs are Probabilistic Graphical Models (PGM) that provide a natural language to describe many optimization problems in robotics and SLAM\footnote{Simultaneous Localization and Mapping} \cite{dellaert2021factor}. In this formulation, the IMU constrains the motion in between keyframes based on linearization-point independent energy factors formulated using the kinematic measurement model of an IMU \cite{Forster2017}:
\begin{equation}
\begin{aligned}
\mathbf{R_{n+1}} &= \mathbf{R_{n}}\exp((\mathbf{\tilde{\boldsymbol{\omega}}_n}-\mathbf{b_g}-\boldsymbol{\eta_g})^{\wedge}\Delta t)\\
\mathbf{v_{n+1}} &= \mathbf{v_n} +\mathbf{g}\Delta t+\mathbf{R_n}(\mathbf{\tilde{a}_n}-\mathbf{b_a}-\boldsymbol{\eta_a})\\
\mathbf{p_{n+1}} &= \mathbf{p_n} + \mathbf{v_n}\Delta t + \frac{1}{2}\mathbf{g}\Delta t^2 +\mathbf{R_n}(\mathbf{\tilde{a}_n}-\mathbf{b_a}-\boldsymbol{\eta_a})\Delta t^2
\end{aligned}
\label{eq:imu_physics}
\end{equation}
where $\mathbf{R_n}$, $\mathbf{v_n}$, and $\mathbf{p_n}$ are orientation, velocity and position of the sensor with respect to the world frame, and $\Delta t$ is the sampling time of the IMU. The angular velocity and acceleration measurements $\boldsymbol{\tilde{\omega}_n}, \boldsymbol{\tilde{a}_n}$ from the gyroscope and accelerator are contaminated with additive Gaussian noise $\boldsymbol{\eta_g},\boldsymbol{\eta_a}$ and random walk bias terms $\boldsymbol{b_g},\boldsymbol{b_a}$. Furthermore, the $\exp(\cdot)$ is the SO3 exponential map that links lie algebra $\mathfrak{so}(3)$ vectors to their corresponding SO3 matrices.

Batches of IMU measurements between two $\mathrm{i}$ and $\mathrm{j}$ keyframes may be compressed into motion constraints by integrating the above equations. Nevertheless, the outcome of this integration changes with the initial condition at time $i$. Multiplying both sides of this integration by $\mathrm{R_i}$ breaks this dependence and yields the following initial conditions independent constraints \cite{Forster2017}:

\begin{equation}
\label{eq:pi_features}
\begin{aligned}
\Delta \mathbf{R}_{i j} & \triangleq \mathbf{R}_{i}^{\top} \mathbf{R}_{j}=\prod_{k=i}^{\jmath-1} \exp\left(\left(\tilde{\boldsymbol{\omega}}_{k}-\mathbf{b}_{k}^{g}-\boldsymbol{\eta}_{k}^{g}\right) \Delta t\right) \\
\Delta \mathbf{v}_{i j} & \triangleq \mathbf{R}_{i}^{\top}\left(\mathbf{v}_{j}-\mathbf{v}_{i}-\mathbf{g} \Delta t_{i j}\right)\\&=\sum_{k=i}^{j-1} \Delta \mathbf{R}_{i k}\left(\tilde{\mathbf{a}}_{k}-\mathbf{b}_{k}^{a}-\boldsymbol{\eta}_{k}^{a}\right) \Delta t \\
\Delta \mathbf{p}_{i j} & \triangleq \mathbf{R}_{i}^{\top}\left(\mathbf{p}_{j}-\mathbf{p}_{i}-\mathbf{v}_{i} \Delta t_{i j}-\frac{1}{2} \sum_{k=i}^{j-1} \mathbf{g} \Delta t^{2}\right) \\
&=\sum_{k=i}^{j-1}\left[\Delta \mathbf{v}_{i k} \Delta t+\frac{1}{2} \Delta \mathbf{R}_{i k}\left(\tilde{\mathbf{a}}_{k}-\mathbf{b}_{k}^{a}-\boldsymbol{\eta}_{k}^{a}\right) \Delta t^{2}\right]
\end{aligned}
\end{equation}

In this paper, we exploit these on-manifold integrated motion constraints as efficient motion features for learning-based models. In other words, rather than feeding a recurrent neural network (RNN) with the raw and high-frequency IMU data, we use preintegration in order to reduce the temporal dimension of the inputs, which is the root of the achieved efficiency. Moreover, the reduced temporality and the induced geometrical bias in formulating this input leads to improved predictive performance as shown in our experiments. 

It is important to note that in this work, we disregard the biases in Eq. \ref{eq:pi_features} by assigning them with random values during the training as a data augmentation technique.

\subsection{The Baseline Model}
\subsubsection{The Network Architecture}
Drawing inspiration from the double integration nature of the IMU kinematics model, we have adopted a two-layer bidirectional LSTM architecture as our baseline neural network. As illustrated in Fig. \ref{base_model}, this model is comprised of two layers with 32 and 128 hidden units. The features extracted by the second LSTM layer are then linearly mapped into 6-DOF $\mathfrak{se}(3)$ odometry labels. These $\mathfrak{se}(3)$ vectors are the logarithmic maps of the expected $SE(3)$ odometry transformations between poses $\boldsymbol{T}_{i-1}$ and $\boldsymbol{T}_{i}$:

\begin{equation}\begin{aligned}
\boldsymbol{\xi}_i=\log{(\boldsymbol{T}_{i-1}^{-1}\boldsymbol{T}_{i})}^\vee \in \mathbb{R}^6
\end{aligned}\end{equation}

Our motivation for bi-directionality is the ability of this model to attend to both future and the past samples within a window of IMU measurements, and we have chosen the larger 128-unit hidden size for the second layer to facilitate better $\mathfrak{se}(3)$ regression. Overall, the baseline model has $210K$ parameters.  

\subsubsection{Loss Formulation}
We adopt the formulation presented in \cite{valentin2018dpcnet} and formulate the training loss as the weighted sum of squares of the geodesic distances between the ground truth odometry labels, $\Delta\boldsymbol{T^*}_{i-1,i}=\boldsymbol{T^*}_{i-1}^{-1}\boldsymbol{T^*}_{i}$, and the network's outputs as follows:
\begin{equation}\begin{aligned}
L&=\sum_{k=1}^{N}\boldsymbol{g}_i^T\boldsymbol{\Sigma}^{-1} \boldsymbol{g}_i\\
\boldsymbol{g}_i&=log(\boldsymbol{\Delta T^*}_{i,i+1}^{-1}\exp{(\boldsymbol{\xi}_i^{\wedge})})^\vee
\end{aligned}\end{equation}
Where $(.)^\wedge$ and $(.)^\vee$ operators respectively transform the $\mathfrak{se}(3)$ vectors into their skew symmetric matrix form and vise versa. The $\boldsymbol{\Sigma}$ parameter in the above equation is an empirical covariance matrix computed using the training data based on the following equations:

\begin{equation}\begin{aligned}
\boldsymbol{\Sigma}&=\frac{1}{N-1} \sum_{i=1}^{N}\left(\boldsymbol{\xi}_{i}^{*}-\overline{\boldsymbol{\xi}^{*}}\right)\left(\boldsymbol{\xi}_{i}^{*}-\overline{\boldsymbol{\xi}^{*}}\right)^{T}\\
\overline{\boldsymbol{\xi}^{*}} &\triangleq \frac{1}{N} \sum_{i=1}^{N} \boldsymbol{\xi}_{i}^{*}
\end{aligned}
\label{eq:emperical_covar}
\end{equation}
where $N$ is the number of labels in the training dataset, and $\boldsymbol{\xi}^{*}_i=\log{(\Delta\boldsymbol{T}_{i-1,i})}^\vee$. The computed covariance in the above equation balances the relative importance of each motion axis and stabilizes the training. 

\section{Experiments and Studies}
\label{experiments}
\subsection{Datasets} 
We evaluate our approach on two different application domains: (1) autonomous driving, and (2) pedestrian odometry. 
For \emph{autonomous driving}, the KITTI odometry dataset \cite{Geiger2012CVPR} has been chosen. This dataset is recorded using a car equipped with vision, Lidar, and RTK-GPS+IMU units traveling around urban and countryside environments. The IMU data is recorded at a rate of 100 Hz, and the ground truth is provided at 10 Hz through a batch maximum likelihood estimation based on Lidar and GPS data. 

For the \emph{pedestrian odometry} application, the handheld and trolley domains of the OxfordIO dataset \cite{Chen2018OxIODTD} have been chosen. OxfordIO dataset contains the IMU readings from a smartphone held in various configurations and carried by a user while undergoing different motion patterns. The ground truth for this dataset is provided using a Vicon motion capture system with millimeter-level accuracy.

\subsection{Baseline Implementation}
We implemented our model using PyTorch and PytorchLightning frameworks and ran the training process on an Nvidia T4 GPU. Adam optimizer with a learning rate of $0.001$ is used to train the models. On average, the convergence happened after 120 epochs. To avoid overfitting, dropout layers have been used after each LSTM layer to zero out $25\%$ of the neuron activities at each iteration. Our implementations will be open sourced\footnote{\url{https://github.com/Rooholla-KhorramBakht/pi_net}}.
\subsection{Baseline Results}
\subsubsection{Autonomous Driving} In this section we train three models. As the base of comparison, we train our baseline model using raw IMU data over 1280 IMU samples which corresponds to 12.8 seconds of IMU measurements. Furthermore, to compare preintegration with a naive Euclidean averaging, we averaged each 10 IMU sample into one measurement and trained a model with 128 such measurements. Finally, our main experiment trains the baseline model on preintegrated features each computed over 10 IMU samples, which leads to 128 input features. Our motivation for choosing an integration factor of 10 has been the 10 Hz ground-truth frequency of the KTIIT dataset as opposed to the 100 Hz IMU samples. 

The test/train splitting policy for the KITTI dataset follows the common practice in data-driven VIO literature \cite{chen2019selectivevio}, which takes sequences 05, 07, and 10 for testing, and sequences \{00-10\}-\{05,07,10,03\} for training. We report our results as relative translation and rotation errors defined by the KITTI benchmark \cite{Geiger2012CVPR}. These relative errors are defined as the averaged position/orientation errors within all possible sub-sequences of lengths $100m,...,800m$. 
\begin{table}[t!]
	\centering
	\caption{Relative odometry translation ($\%$) and orientation ($deg/100m$) errors on the KITTI dataset.}
	\label{table:kittitable}
	\begin{tabular}{c||c|c||c|c||c|c} 
		\toprule
		\multirow{2}{*}{\begin{tabular}[c]{@{}c@{}}test\\seq.\end{tabular}} & \multicolumn{2}{c||}{\begin{tabular}[c]{@{}c@{}}Raw IMU \end{tabular}}                                            & \multicolumn{2}{c||}{\begin{tabular}[c]{@{}c@{}}Averaged IMU\end{tabular}}                                              & \multicolumn{2}{c}{\begin{tabular}[c]{@{}c@{}}\textbf{PI-Features} \textbf{(ours)}\end{tabular}}                            \\
		& \begin{tabular}[c]{@{}c@{}}$t_{rel}$\end{tabular} & \begin{tabular}[c]{@{}c@{}}$r_{rel}$\end{tabular} & \begin{tabular}[c]{@{}c@{}}$t_{rel}$\end{tabular} & \begin{tabular}[c]{@{}c@{}}$r_{rel}$\end{tabular} & \begin{tabular}[c]{@{}c@{}}$t_{rel}$\end{tabular} & \begin{tabular}[c]{@{}c@{}}$r_{rel}$\end{tabular}  \\ 
		\hline
		05                                                                  & 10.95                                                     & 2.06                                                            & 17.02                                                     & 2.73                                                            & \textbf{4.75}                                             & \textbf{0.85}                                                    \\
		07                                                                  & 16.07                                                     & 4.42                                                            & 14.04                                                     & 2.04                                                            & \textbf{9.06}                                             & \textbf{1.56}                                                    \\
		10                                                                  & 7.57                                                      & 1.52                                                            & 9.03                                                      & 2.41                                                            & \textbf{5.26}                                             & \textbf{0.75}                                                    \\ 
		\hline
		avg.                                                                & 11.53                                                     & 2.66                                                            & 13.36                                                     & 2.38                                                            & \textbf{6.35}                                             & \textbf{1.05}                                                    \\
		\bottomrule
	\end{tabular}
\end{table}

\begin{figure*}
	\begin{subfigure}[t]{0.32\textwidth}
		\includegraphics[width=1\textwidth]{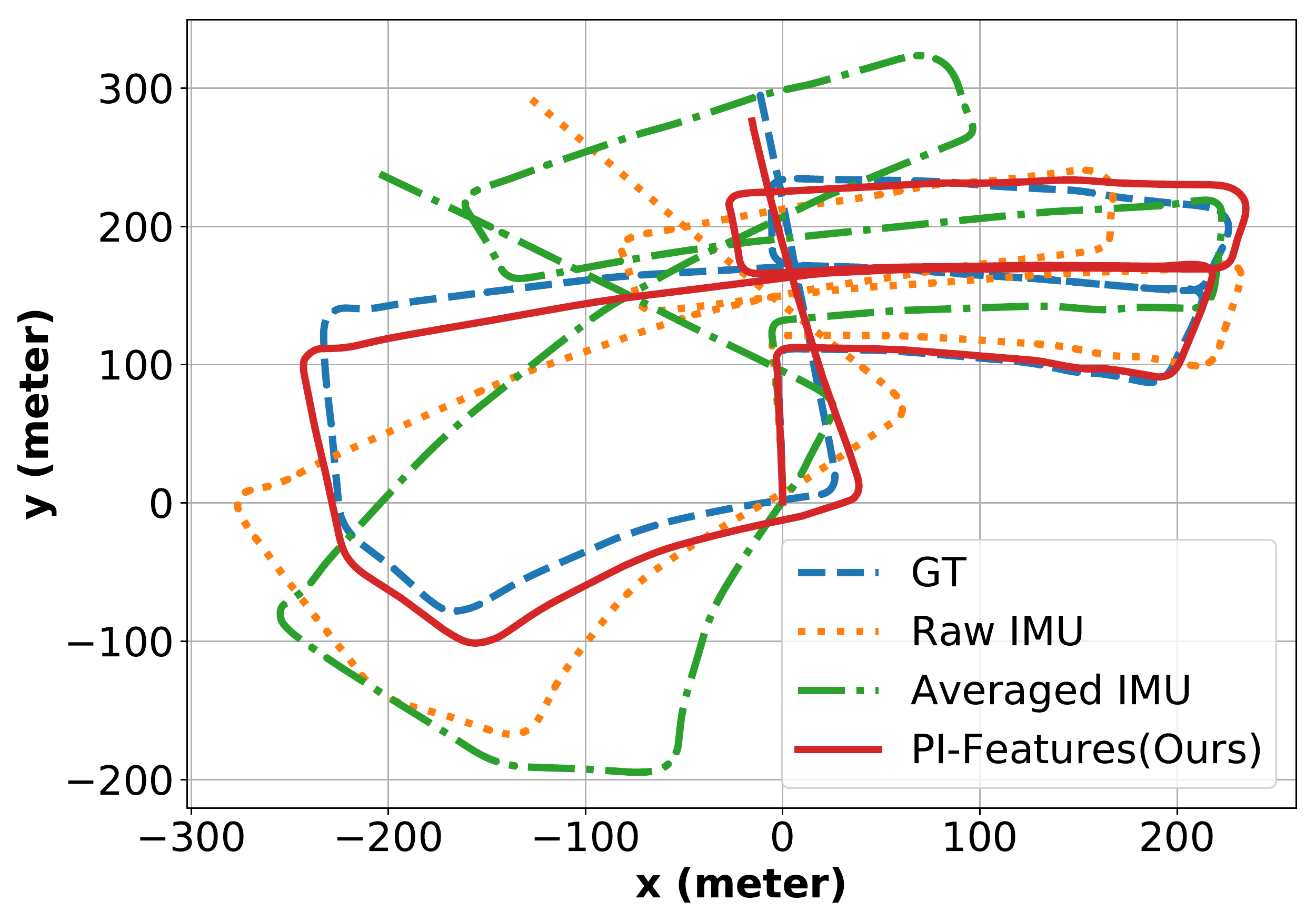}
		\caption{Seq. 05}
		\label{fig:x1s}
	\end{subfigure}
	\hfill
	\begin{subfigure}[t]{0.32\textwidth}
		\includegraphics[width=1\textwidth]{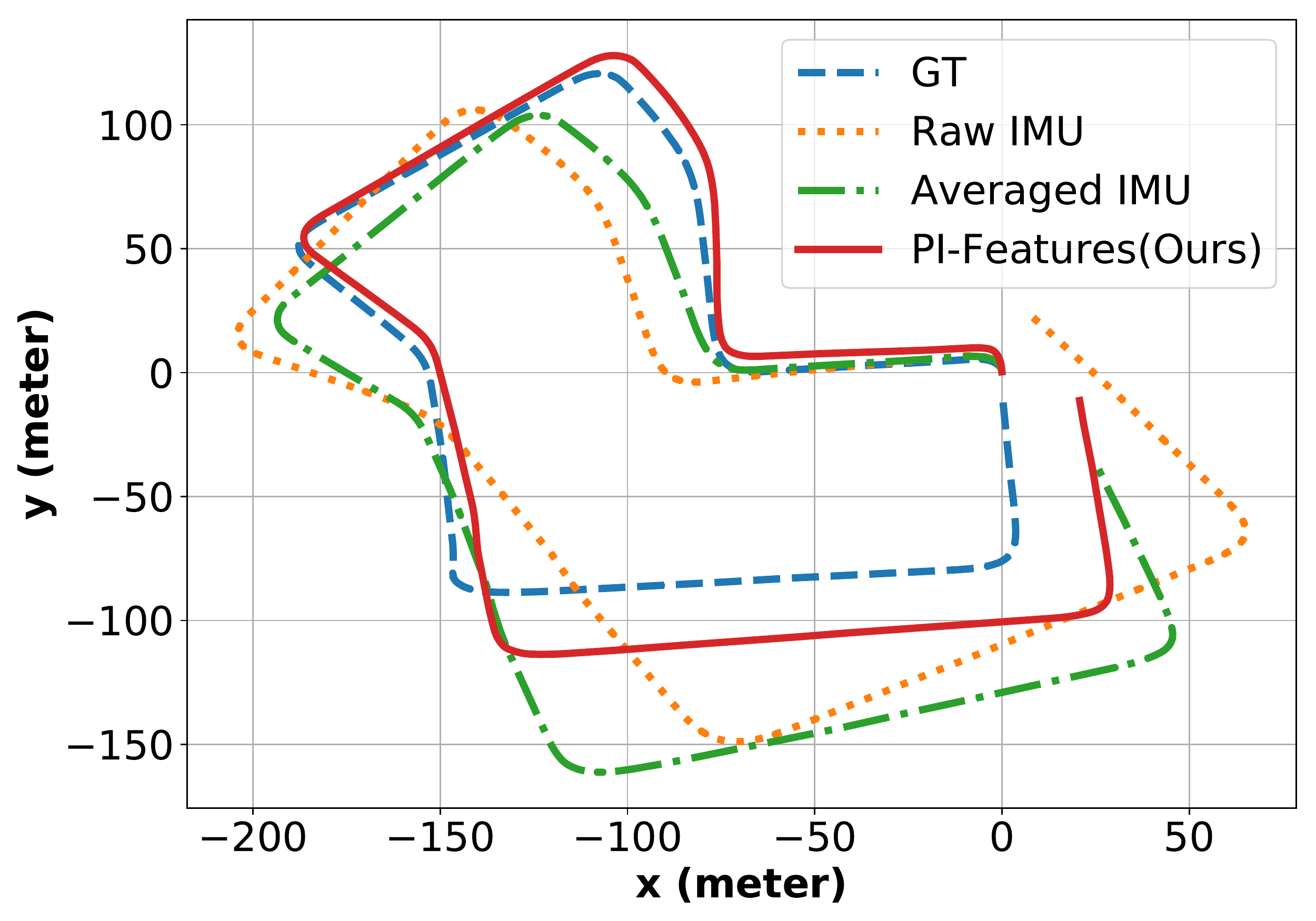}
		\caption{Seq. 07}
		\label{fig:x2}
	\end{subfigure}
	\hfill
	\begin{subfigure}[t]{0.32\textwidth}
		\includegraphics[width=1\textwidth]{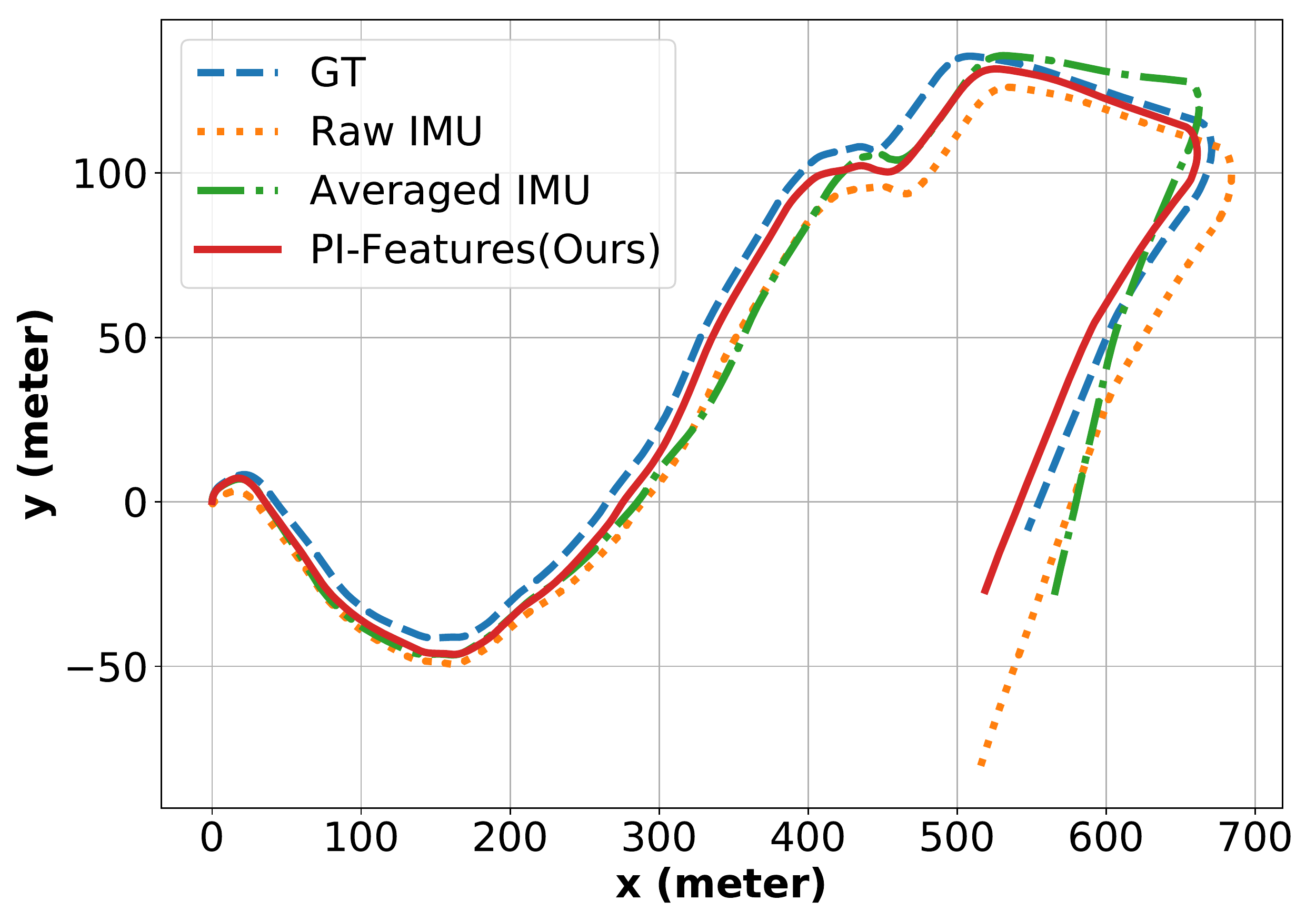}
		\caption{Seq. 10}
		\label{fig:x3}
	\end{subfigure}
	\caption{Qualitative comparison between the baseline model with raw, averaged, and preintegrated IMU inputs on KITTI datasets. This figure is best viewed in the color version.}
	\label{fig:kittifigs}
\end{figure*}

Table \ref{table:kittitable} represents the overall comparison between the results of our three experiments. As it can be seen from the first and second columns of the table, the average translation and orientation errors of averaged measurements is the lowest and as shown in the third column, the baseline model trained using the preintegrated features consistently achieves the best results compared to both averaged and raw inputs. This performance improvement is also qualitatively evident from Fig. \ref{fig:kittifigs}. As shown in this figure, the models trained using the preintegrated IMU features exhibit better orientation stability which in turn leads to better translational accuracy. 

\begin{table*}[t!]
\centering
\caption{Relative translation errors on OxfordIO dataset (handheld/trolley-data\emph{\#}-sequence\emph{\#})}
\label{tab:oxfordIO}
\begin{tabular}{c|c|c|c|c|c|c|c|c|c}
\hline
Method                                              & h-d1-s2         & h-d1-s5         & h-d1-s6         & h-d3-s1         & h-d4-s1         & h-d4-s3         & h-d5-s1        & t\_d2\_s6                        & average         \\ \hline
\textbf{PI Features (Ours)} ($\%$) & $\mathbf{5.52}$ & $\mathbf{2.66}$ & $\mathbf{2.98}$ & $\mathbf{3.64}$ & $\mathbf{4.16}$ & $\mathbf{3.27}$ & $\mathbf{3.3}$ & $7.61$                           & $\mathbf{4.14}$ \\
Raw IMU ($\%$)                             & $6.5$           & $3.33$          & $3.12$          & $4.45$          & $4.32$          & $4.6$           & $3.64$         & $\mathbf{4.87}$ & $4.35$          \\ \hline
\end{tabular}
\end{table*}


We believe that the geometrical inductive bias in calculating the preintegrated features and the reduced burden of memorizing long sequences of raw measurements contribute to this performance improvement when using preintegrated features. Furthermore, as shown in the second column of the table, a naive averaging of the IMU measurements leads to information loss and loss of performance compared to a model trained using raw IMU inputs. 

\subsubsection{Pedestrian Odometry} We also investigate the effectiveness of our approach for the domain of pedestrian odometry using the OxfordIO dataset. Since naive averaging was shown to be ineffective for deep inertial odometry in the previous section, this section considers only the model trained using the raw IMU inputs as the baseline. 

As a metric, we integrate the 6-DoF odometry outputs from both models for batches of 200 IMU measurements, which translates to 2 seconds of walking. The two predictions are then compared against the ground truth to compute the error for each model. We then divide these errors by the displacement length to normalize them.

The experiments in the OxfordIO dataset have been carried out for the trolley and handheld motion domains. The relative translation errors are described in Table \ref{tab:oxfordIO}. As it can be seen from comparing errors in the first and second row of Table \ref{tab:oxfordIO}, a model trained with preintegrated IMU features as input consistently surpasses the baseline trained using the raw IMU measurements. 

The only exception is for the trolley domain. Our intuition is that the model trained using the raw data on the trolley domain enjoys the unattenuated high-frequency contents induced by the vibrations of the wheel.  This spectral modality helps the model to distinguish various motion states of the trolley and is filtered out by the low pass nature of the integration in the computation of the preintegrated features. As a future research direction, we aim to investigate the extension of preintegrated features with spectral information.

\subsection{Embedded Implementation}
The baseline architectures in the previous subsection were executed on an NVIDIA GTX1050 GPU and required $1 ms$ to process $20\times9$ PI features and $3.7 ms$ for its corresponding $200\times6$ raw IMU samples (two seconds of data in our datasets). This substantial performance improvement is due to the compressed temporal dimension of the IMU thus fewer LSTM recursions. In this section, we aim to show that IMU preintegration can enable deep inertial odometry even for highly constrained platforms such as microcontrollers. To this aim, we reproduce the IO-Net formulation of deep inertial odometry \cite{Chen2020}. Specifically, we train a small Convolutional Neural Network (CNN)  that takes a 2D $9\times T$ feature map created by stacking $T$ preintegrated IMU features as input and predicts the corresponding translation and heading change in polar coordinate. For comparison purposes, we also construct a bi-LSTM model with the same output format. Two versions of this baseline are then trained using the preintegrated features and raw IMU measurements as inputs. 

\subsubsection{Architecture} 
The architecture of the LSTM baseline is similar to the model shown in Fig. \ref{base_model}. The only difference is that the predictions are taken from the feature map of the last timestamp, and the number of hidden units are chosen to be 128 and 256 for the first and second layers to match the values reported in \cite{Chen2020}. The feature map corresponding to the last timestamp is then linearly mapped to a $\Delta \Phi$ and $\Delta L$ polar odometry prediction. Similar to the design choice of the IO-Net implementation, we break the raw IMU data into separate windows of 200 samples (2 seconds) and process each of them individually to yield the odometry output
$(\Delta L,\Delta \Phi)$ in polar coordinates. Each odometry output updates the subject's position based on the following kinematic model:
\begin{equation}
\begin{cases}
\Phi_{n+1}&=\Phi_{n-1}+\Delta \Phi\\
X_{n}&=X_{n-1}+\Delta L cos(\Phi_{n})\\
Y_{n}&=Y_{n-1}+\Delta L sin(\Phi_{n})
\end{cases}
\label{eq:integration_formula}
\end{equation}

Based on the above equation, the old heading $\Phi_n$ is updated
using the $\Delta \Phi$ odometry output. The updated $\Phi_{n+1}$ is
then used alongside the odometry output $\Delta L$ to update the
position $P_n=[X_n, Y_n]^T$.

\begin{figure}[b]
	\centering
	\includegraphics[scale=1.1]{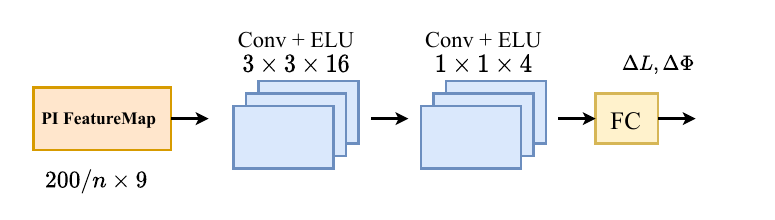}
	\caption{ The proposed CNN model for the embedded implementation of the
		IO-Net pedestrian odometry. In the figure, $n$ is the integration length and in this paper is set to 10.}
	\label{cnn_model}
\end{figure}

A smaller feedforward CNN model has been chosen for the microcontroller implementation. As shown in Fig. \ref{cnn_model}, this model takes the stacked PI features as input and passes them through two CNN layers, one with a kernel size of $3 \times 3 \times 16$ and the other with a kernel size of $1 \times 1 \times 4$. The extracted feature is eventually mapped into two scalar outputs corresponding to the $\Delta \Phi$ and $\Delta L$ polar odometry predictions.

It is important to note that the temporally compressed preintegrated
features enable shallow CNNs to cover a much larger temporal receptive field of the motion signal. As a result, one may drastically shrink the model dimension and kernel sizes which facilitate the embedded implementation of the model.

\subsubsection{Training and Metrics} We train the model using the mean squared difference between the expected and predicted polar values which is formulated as follows:

\begin{equation}
L=\sum_{i=1}^{N}||\Delta L_i - \Delta \hat{L}_i||_2^2+\beta ||\Delta \Phi_i - \Delta \hat{\Phi}_i||_2^2
\label{eq:loss_funtion}
\end{equation}
where $N$ is the number of training sequences in the batch, $\Delta L_i,\Delta
Phi_i$ are the network's outputs, and $\Delta \hat{L_i},\Delta
\hat{\Phi_i}$ are the ground-truth labels. Furthermore, the $\beta$
hyperparameter in the above equation provides a balance between the
orientation and translation components of the loss. This paper sets
this parameter based on the average norm of the orientation and
translation losses on a randomly initialized model such that the two
terms would contribute equally in the backpropagation.

$\Delta \hat{\Phi_i}$ heading label is the projection of the
orientation difference between the first and the last sample in the
window, and is calculated as follows:
\begin{equation}
\Delta \hat{\Phi_i}=log(\boldsymbol{R_i}^T\boldsymbol{R_{i+199}})^\vee
\end{equation}
where $\log(.)$ function in the above equation is the $SO3$
logarithmic map and $R_i$ represents the ground-truth label for the
$i^{th}$ sample. Furthermore, $\Delta \hat{L_i}$ is
calculated as the Euclidean distance between the x-y projections of
the first and the last position within a window:
\begin{equation}
\Delta \hat{L_i}=\sqrt{(X_i-X_{i+199})^2+(Y_i-Y_{i+199})^2}
\end{equation}
The Root Mean Square Error (RMSE) metric has also been adopted for comparing the results quantitatively. Instead of comparing the integrated paths, we opt to compare the odometry outputs since it provides a more direct comparison.

\subsubsection{Performance of The Trained Models}
\begin{table*}[t]
    \captionsetup{justification=centering,margin=2cm}
    \caption{RMS error of CNN and baseline IO-Net models trained using preintegrated and raw IMU measurements.}\begin{tabular}{cccc}
        \toprule
        \multicolumn{1}{l}{} & \begin{tabular}[c]{@{}c@{}}\textcolor[rgb]{0.2,0.2,0.2}{Base Model}\\\textcolor[rgb]{0.2,0.2,0.2}{$(\Delta L (m/S),\Delta \Phi (rad/S))$}\\\textcolor[rgb]{0.2,0.2,0.2}{}\end{tabular} & \begin{tabular}[c]{@{}c@{}}\textcolor[rgb]{0.2,0.2,0.2}{Model with Preintegrated Input }\\\textcolor[rgb]{0.2,0.2,0.2}{$(\Delta L (m/S),\Delta \Phi (rad/S))$}\end{tabular} & \begin{tabular}[c]{@{}c@{}}\textcolor[rgb]{0.2,0.2,0.2}{Embedded CNN Model }\\\textcolor[rgb]{0.2,0.2,0.2}{$(\Delta L (m/S),\Delta \Phi (rad/S))$}\end{tabular}  \\
        \hline
        Slow Walking         & $(0.18,0.013)$                                                                                                                                                                             & $(\mathbf{0.017,0.013})$                                                                                                                                                                        & $(0.029,0.013)$                                                                                                                                                   \\
        Running              & $({0.125}, {0.255})$                                                                                                                                                                             & $\mathbf{(0.1,0.216)}$                                                                                                                                                                        & $(0.115,0.26)$                                                                                                                                                   \\
        Pocket               & $(0.0141,0.025)$                                                                                                                                                                             & $\mathbf{(0.014,0.025)}$                                                                                                                                                                        & $(0.021,0.027)$                                                                                                                                                   \\
        Handbag              & $(0.041,0.0261)$                                                                                                                                                                          & $(\mathbf{0.032},\mathbf{0.081})$                                                                                                                                                                        & $(0.048,0.081)$                                                                                                                                                   \\
        \bottomrule
    \end{tabular}

    \label{table:baseline_vs_pi_vs_cnn}
\end{table*}

Table \ref{table:baseline_vs_pi_vs_cnn} presents the model's performances on slow walking, running, phone in pocket, and phone in the handbag domains. The reported numbers are the average rate of change for the heading and stride predictions, $(\Delta L/\Delta T,\Delta \Phi/\Delta T)$. The first and the second columns present the baseline LSTM model trained using the raw inputs, and the second column represents the result trained using the preintegrated features. 

Furthermore, The third column presents the performance of the small CNN model. As verified in the previous sections, with preintegration, the heading and stride predictive performance is improved compared to the model trained with raw inputs. On the other hand, we can see that the CNN model performs reasonably well even though its complexity is highly reduced compared to the stacked LSTM baseline. Despite this simplicity, it is interesting to note that the CNN model trained on preintegrated inputs performs on par with the baseline IO-Net trained using raw measurements.

\subsubsection{Implementation}
The embedded implementation of the CNN model is carried out using the
X-Cube-AI framework \cite{STMicroelectronics2020}, and the
preintegration processing has been implemented using the highly
optimized ARM CMSIS-DSP math library. While we could have quantized
the Keras model for improving its performance, we used the
floating-point version due to its generality and faster
implementation. On the lowest level of the X-Cube-AI framework, a
run-time utilizes the ARM CMSIS-NN/DSP libraries \cite{ARM2020} to
run the converted Keras network efficiently. The user application
may then load the run-time with the appropriate network for
inference. In this paper, the target platform is an ARM
STM32F407ZET microcontroller with 198KB of RAM and 512KB of flash
memory, and with a Cortex-M4 CPU running at 168Mhz.
\begin{table*}
    \centering
    \captionsetup{justification=centering,margin=2.5cm}
    \caption{Resource consumption and performance on an \emph{STM32F407ZET} microcontroller with $512 KBi$ of Flash and $192 KBi$ of RAM running at a clock frequency of $168 MHz$.}
    \begin{tabular}{cccc}
        \toprule
        \textbf{Process}            & \textbf{RAM Usage (KB)} & \textbf{ROM Usage (KB)}               & \textbf{Execution Time (ms)}  \\
        \hline
        Preintegration (per vector) & \multicolumn{1}{l}{4 (Stack Size)}     & \multicolumn{1}{l}{~0 (No Parameters)} & 0.2                           \\
        \hline
        Model Inference             & 12.28                    & 92.55                                  & 5                             \\
        \bottomrule
    \end{tabular}
    \label{table:embedded_implementation}
\end{table*}
Table \ref{table:embedded_implementation} presents the
implementational details of the CNN model on this microcontroller. As seen in the table, calculating each
preintegration vector takes about 0.2ms while each model inference requires
only 5ms of computational time. Assuming a stride of 10 (running the
inference after each 10 IMU measurements), each inference requires
calculating one preintegrated feature and running one model
inference, which in total takes 5.2ms. Since this computation is
required every 100 milliseconds, there will
be plenty of CPU time left for other user-specific tasks. 

Furthermore, the model's memory consumption is reasonably low and is around $92.55/512KB$ for flash and $12.28/192KB$ for RAM which leaves plenty
of room for other user-specific data and code. Furthermore, the successful processing of the preintegration vectors in the microcontroller required increasing the stack to 4KB which is reported here as the approximate required RAM for performing preintegration.
\section{Conclusions and Future Works}
\label{conclusions}
This paper proposed a computationally efficient preprocessing method for raw IMU measurements that utilizes the manifold structure of the IMU model to compress long sequences of raw measurements into short and effective motion
representations for deep inertial odometry. This compression improves the
computational efficiency of deep motion estimation in embedded applications. In addition to this added efficiency, this paper showed improved odometry performance based on preintegrated features through experiments carried out in two distinct domains of pedestrian inertial odometry and autonomous vehicles. Finally, the efficiency of our approach was demonstrated through the formulation of a well-known pedestrian odometry model based on preintegrated features and its implementation on an embedded microcontroller with
highly-restricted resources. Prominent directions for future research are
applying preintegration for other applications such as human
activity recognition and visual-inertial odometry. Furthermore, the proposed PI feature may be enriched using acoustic features such as FFTs to encode the vibration modality alongside the preintegrated motion vectors, which may provide further improvements in terms of accuracy and robustness.

\bibliographystyle{ACM-Reference-Format}
\bibliography{References}
\end{document}